\documentclass[letterpaper]{article} 
\usepackage{aaai23}  
\usepackage{times}  
\usepackage{helvet}  
\usepackage{courier}  
\usepackage[hyphens]{url}  
\usepackage{graphicx} 
\usepackage{natbib}  
\usepackage{caption} 
\usepackage{algorithm}
\usepackage{algorithmic}

\usepackage{booktabs}       
\usepackage{xcolor}         
\usepackage{amsmath,amssymb}
\usepackage{textcomp}
\usepackage{multirow}
\usepackage{graphicx}
\usepackage{amsthm}
\newtheorem{thm}{Theorem}

\newtheorem{remark}[thm]{Remark}
\newcommand{\tblue}[1]{\textcolor{black}{#1}}

\newcommand{\G}{\mathcal{G}}
\newcommand{\F}{\mathcal{F}}
\newcommand{\K}{\mathcal{K}}

\usepackage{newfloat}
\usepackage{listings}
\DeclareCaptionStyle{ruled}{labelfont=normalfont,labelsep=colon,strut=off} 
\floatstyle{ruled}
\newfloat{listing}{tb}{lst}{}
\floatname{listing}{Listing}
\title{Rule Induction in Knowledge Graphs Using Linear Programming}
\author{
    Sanjeeb Dash, Jo\~ao Gon\c calves
}
\affiliations{
    IBM Research\\

    Yorktown Heights, New York, USA\\
    \texttt{\{sanjeebd, jpgoncal\}@us.ibm.com}
}

\begin{document}

\maketitle

\begin{abstract}
We present a simple linear programming (LP) based method to learn compact and interpretable sets of rules encoding the facts in a knowledge graph (KG) and use these rules to solve the KG completion problem. Our LP model chooses a set of rules of bounded complexity from a list of candidate first-order logic rules and assigns weights to them. The complexity bound is enforced via explicit constraints. We combine simple rule generation heuristics with our rule selection LP to obtain predictions with accuracy comparable to state-of-the-art codes, even while generating much more compact rule sets. Furthermore, when we take as input rules generated by other codes, we often improve interpretability by reducing the number of chosen rules, while maintaining accuracy.
\end{abstract}

\section{Introduction}
\label{submission}

Knowledge graphs (KG) represent a collection of known facts via labeled directed edges.
A {\em fact}  is a triplet of the form  
$(a, r, b)$, where $a$ and $b$ are nodes representing {\em entities}, and $r$ is a binary relation labeling a directed edge from $a$ to $b$ indicating that $r(a,b)$ is true. 
Practical knowledge graphs are often incomplete (they do not contain all true representable facts).
Knowledge graph completion (KGC) involves using known facts in a KG to infer additional (missing) facts. Other common tasks for extracting implied information from KGs are triple classification, entity recognition, and relation prediction.
See the survey by \citet{kg-survey}.

One approach for KGC is
to learn first-order logic (FOL) rules that approximately encode known facts (in practice, KGs have inconsistencies).
Consider a KG where the nodes correspond to distinct cities, states, and countries and the relations are one of $\{${\it capital\_of}, {\it shares\_border\_with}, {\it part\_of}$\}$. Suppose we learn a ``rule''  {\it capital\_of(X,Y) and part\_of(Y,Z)} $\rightarrow$ {\it part\_of(X,Z)} of {\em length} two, where {\it X, Y, Z} are variables. When we apply this rule to entities {\it P} and {\it Q}, if there exists a third entity {\it R} such that {\it (P, capital\_of,R)} and {\it (R,part\_of,Q)} are facts in the graph then we infer that {\it P} is a part of {\it Q}. 
KGC deals with finding answers to queries of the form {\it ( P, part\_of, ?)}.
We focus on finding multiple FOL rules of the type above for this task along with rule weights, where larger weights indicate more important rules. 

Learning logic rules is a well-studied area. In a paper \citep{lao-cohen} on path ranking algorithms and another \citep{rich-dom} on Markov logic networks, candidate logic rules are obtained via relational path enumeration, and rule weights are calculated. \citet{yang17} use neural logic programming to simultaneously obtain rules and rule weights. In \cite{RNNLogic}, rules and rule weights are computed in sequence, but a feedback loop from the latter learning problem to the former is added. \tblue{The use of recursive neural networks (RNN) to learn rules is common nowadays, though traditional rule-mining approaches can be effective \citep{anyburl}.}

Embedding-based methods for KGC consist of representing nodes by vectors, and relations by vector transformations that are consistent with the KG facts.
They exhibit better scaling with KG size, and yield more accurate predictions. See the surveys by \citet{kg-survey} and \citet{emb-survey}. However, they are less interpretable than logical rules.

Besides predictive accuracy, \citet{michie} identified {\em comprehensibility} as an important property of a machine learning (ML) method and proposed the {\em strong criterion} and {\em ultra-strong criterion}. The former is satisfied if the ML system generates an explicit symbolic representation of its hypotheses, and the latter is satisfied if humans can understand and effectively apply such symbolic representations.
There is much current interest in interpretable ML systems; see \cite{interp-cynthia} and \cite{iai}. In socially sensitive applications of ML, e.g., in the criminal justice system, interpretability is viewed as an essential feature to allow audits for fairness and bias avoidance \cite{bias}. 
Experimental results in \cite{mug1, mug2} show that for logic programs {\em inspection time} (the time taken by humans to study and understand the program before applying it) is negatively correlated with human predictive performance. Though rule-based methods satisfy the strong criterion of Michie, many such methods generate many rules which take a lot of time to inspect, leading to low comprehensibility and interpretability. 
The tradeoff between accuracy and rule simplicity (or compactness) is well-studied in the setting of classification \cite{dash2018boolean,rudin17} where it is shown that one can obtain interpretable rule sets without significantly sacrificing accuracy.

Inspired by the above classification methods, we focus on learning compact sets of entity independent FOL rules for KGC (as KGC has {\it no negatively} labeled examples, one cannot apply classification techniques directly).
We combine rule enumeration with linear programming (LP) and avoid solving difficult nonconvex optimizaton models inherent in training RNNs, though the difficulty is transferred to rule enumeration. 
We describe an LP formulation with one variable per candidate FOL rule and its associated weight. Nonzero variable values in the solution correspond to the weights of chosen rules.
Our output scoring function is a linear combination of rule scores.
Linear combinations are also used in NeuralLP \citep{yang17}, DRUM \citep{sadeghian19}, and RNNLogic \citep{RNNLogic} though we calculate scores differently.

 To promote interpretability, we  add a constraint limiting the complexity of the chosen rule set. 
We show that initializing our LP with a few rules enumerated via simple heuristics leads to high-quality solutions with a small number of chosen rules for some benchmark KGs; we either obtain better accuracy or more compact rule sets in all cases compared to well-known rule-based methods.
Our algorithm has better scaling with KG size than the above rule-based methods and can scale to YAGO3-10, a large dataset that is difficult for many rule-based methods (though AnyBURL \cite{anyburl} is faster). 
For YAGO3-10, where setting up the LP becomes expensive, we 
use column generation ideas from linear optimization and start off with a small initial set of rules, find the best subset of these and associated weights via the partial LP defined on these rules, and then generate new rules which can best augment the existing set of rules.

We explore taking as input rules generated by other rule-based methods and then choosing the best subset using our LP. In some cases, we improve accuracy even while reducing the number of rules. As compact rules do well in an inductive setting, we compare against GraIL \cite{grail} -- a method that uses subgraph reasoning for KGC -- in such a setting and obtain better performance.


\section{Related Work}


\tblue{The motivation for learning rules for KG reasoning is that they form an explicit symbolic representation of the KG and are amenable to inspection and verification, especially when the number of rules is small. 
}

{\bf Inductive Logic Programming (ILP).} \tblue{In this approach, one takes as input positive and negative examples and learns logic programs that entail all positive examples and none of the negative examples. See \cite{metagol-ai} and \cite{learn-fail}. FOL programs in the form of a collection of chain-like Horn clauses are a popular output format. Negative examples are not available in typical knowledge graphs, and some of the positive examples can be mutually inconsistent. \citet{evans} developed a differential ILP framework for noisy data.}

{\bf Statistical Relational Learning (SRL).} \tblue{SRL aims to learn FOL formulas from data and to quantify their uncertainty. Markov logic, which is a probabilistic extension of FOL, is a popular framework for SRL. In this framework, one learns a set of weighted FOL formulas.}  \citet{kok} use beam search to find a set of FOL rules, and learn rule weights via standard numerical methods. In knowledge graph reasoning, {\em chain-like} rules which correspond to relational paths (and to chain-like Horn clauses) are widely studied.
\tblue{ A recent, bottom-up rule-learning algorithm with excellent predictive performance is AnyBURL. We learn FOL rules corresponding to relational paths and rule weights; our scoring function and learning model/algorithm are different from prior work.}


{\bf Neuro-symbolic methods.}
In NeuralLP rules and rule weights are learned simultaneously by training an appropriate RNN. Further improvements in this paradigm can be found in DRUM. NTP \citep{roch} is yet another neuro-symbolic method.
\tblue{More general rules (than the chain-like rules in NeuralLP) are obtained in \citep{yang-song}, \cite{sen_et_al}, \cite{Sen_Carvalho_Riegel_Gray_2022} along with better scaling behavior.}
Simultaneously solving for rules and rule-weights is difficult, and a natural question is how well the associated optimization problem can be solved, and how scalable such methods are. \tblue{We use an easier-to-solve LP formulation.}

{\bf Reinforcement Learning (RL).} Some recent codes that use RL to search for rules are MINERVA \citep{minerva}, MultiHopKG \citep{multihopkg}, M-Walk \citep{NEURIPS2018_c6f798b8} and DeepPath \citep{xiong}. \tblue{The first three papers use RL to explore relational paths conditioned on a specific query, and use RNNs to encode and construct a graph-walking agent. } 

\tblue{{\bf Rule types/Rule combinations.}  
NLIL \cite{yang-song} goes beyond simple chain-like rules. Subgraphs are used in \cite{grail} to perform reasoning, and not just paths. We generate {\em weighted, entity-independent, chain-like rules} as in NeuralLP. Our scoring function combines rule scores via a linear combination.  For a rule $r$ and a pair of entities $a, b$, the rule score is just 1 if there exists a relational path from $a$ to $b$ following the rule $r$ and 0 otherwise. We use rule weights as a measure of importance (but not as probabilities). For Neural LP, DRUM, RNNLogic and other comparable codes, the scoring functions depend on the set of paths from $a$ to $b$ associated with $r$. AnyBURL uses maximum confidence scores.}

\tblue{{\bf Scalability/Compact Rule sets.} As noted above, many recent papers use RNNs in the process of finding chain-like rules and this can lead to expensive computation times. On the other hand, bottom-up rule-learners such as AnyBURL are much faster. The main focus of our work is obtaining compact rule sets for the sake of interpretability while maintaining scalability via LP models and column generation.
NeuralLP usually returns compact rule sets while AnyBURL returns a large number of rules (and does not prune discovered rules for interpretability), and RNNLogic is somewhere in between (the number of output rules can be controlled).}

\section{Model}

We propose an LP model inspired by LP boosting methods for classification using classical {\em column generation} techniques \citep{dbs,eg2,eg3,dash2018boolean}.
 Our goal is to create a weighted linear combination of first-order logic rules to be used as a scoring function for KGC. In principle, our model has exponentially many variables corresponding to the possible rules. In practice, we initialize the LP with few initial candidate rules. If the solution is satisfactory, we stop, otherwise we use column generation and  generate additional rules that can improve the overall solution.

\noindent {\bf Knowledge graphs:}
Let $V$ be a set of entities, and let $\mathcal{R}$ be a set of $n$ binary relations defined over $V\times V$.  
A {\em knowledge graph } represents a set of {\em facts} $\F \subseteq V \times \mathcal{R} \times V$ as a labeled, directed multigraph $\G$.
Let $\F = \{(t^i, r^i, h^i) : i=1, \ldots |\F|\}$ where $t^i \neq h^i \in V$, and $r^i \in \mathcal{R}$.
The nodes of $\G$ correspond to entities in $\F$ and the edges to facts in $\F$: a fact $(t, r, h)$ in $\F$ corresponds to the directed edge $(t,h)$ in $\G$ labeled by the relation $r$, depicted as $t \overset{r}{\rightarrow} h$. Here $t$ is the {\em tail} of the directed edge, and $h$ is the {\em head}.
Let $E$ stand for the list of directed edges in $\G$. 
In practical KGs missing facts that can be defined over $V$ and $\mathcal{R}$ are not assumed to be incorrect.
The {\em knowledge graph completion} task consists of taking a KG as input and answering a list of queries of the form $(t,r,?)$ and $(?,r,h)$, constructed from facts $(t,r,h)$ in a test set. The query $(t,r,?)$ asks for a head entity $h$ such that $(t,r,h)$ is a fact, given a tail entity $t$ and a relation $r$. A collection of facts $\F$ is divided into a training set $\F_{tr}$, a validation set $\F_{v}$, and a test set $\F_{te}$, the KG $\G$ corresponding to $\F_{tr}$ is constructed and a scoring function is learnt from $\G$ and evaluated on the test set.

\noindent {\bf Goal:} For each relation $r$ in $\G$, we wish to learn a scoring function $f : (t,r,h) \rightarrow \mathbb{R}$ that returns high scores for true facts (in $\mathcal{F}$) and low scores for facts not in $\mathcal{F}$.
To do this, we find a set of {\em closed, chain-like} rules $R_1, \ldots, R_p$ and positive weights $w_1, \ldots w_p$ where each rule $R_i$ has the form
\begin{equation}\label{eq1}
 r_1(X,X_1) \wedge r_2(X_1, X_2) \wedge \cdots \wedge r_l(X_{l-1}, Y) \rightarrow r(X,Y).
\end{equation}
Here $r_1, \ldots, r_l$ are relations in $\G$, and the {\em length} of the rule is $l$.
The interpretation of this rule is that if for some entities (or nodes) $X,Y$ of $\G$ there exist entities $X_1, \ldots X_l$ of $\G$ such that $r_1(X,X_1), r_l(X_{l-1}, Y)$ and $r_j(X_{j-1},X_j)$ are true for $j=2, \ldots l-1$, then $r(X,Y)$ is true.
We refer to the conjunction of relations in (\ref{eq1}) as the clause associated with the rule $R_i$.
Thus each clause $C_i$ is a function from $V \times V$ to $\{0,1\}$, and we define $|C_i|$ to be the number of relations in $C_i$. Clearly, $C_i(X,Y) = 1$ for entities $X,Y$ in $\G$ if and only if there is a {\em relational path} of the form
\[X \overset{r_1}{\rightarrow} X_1 \cdots X_{l-1} \overset{r_l}{\rightarrow} Y.\]
Our learned scoring function for relation $r$ is simply
\begin{equation}\label{pred1}
f_r(X,Y) = \sum_{i=1}^p w_i C_i(X,Y) \mbox{ for all } X,Y \in V.
\end{equation}
We learn weights of the linear scoring function above by solving an LP that rewards high scores (close to 1) to training facts and penalizes positive scores given to missing facts. Given a query $(t,r,?)$ constructed from a fact $(t,r,h)$ from the test set, we calculate $f_r(t,v)$ for every entity $v \in V$, and the {\em rank} of the correct entity $h$ is the position of $h$ in the sorted list of all entities $v$ (sorted by decreasing value of $f_r(t,v)$). We similarly calculate the rank of $t$ for the query $(?,r,h)$. We then compute standard metrics such as MRR (mean reciprocal rank), Hits@1, Hits@3, and Hits@10 \citep{sun} in the filtered setting \cite{bordes}. An issue in rank computation is that multiple
entities (say $e'$ and $e''$) can get the same score for a query and different treatment of equal scores can lead to very different MRR values.
In {\it optimistic} ranking, an entity is given a rank equal to one plus the number of entities with strictly larger score.
We use {\it random break} ranking (an option available in NeuralLP), where ties in scores are broken randomly.

\noindent {\bf New LP model for rule learning for KGC.} 
Let $\K$ denote the set of clauses of possible rules of the form (\ref{eq1}) with maximum rule length $L$.
Clearly, $|\K| = n^L$, where $n$ is the number of relations.
Let $E_r$ be the set of edges in $\G$ labeled by relation $r$, and assume that $|E_r| = m$.
Let the $i$th edge in $E_r$ be $(X_i,Y_i)$. We compute $a_{ik}$ as $a_{ik} = C_k(X_i,Y_i)$: $a_{ik}$ is 1 if and only if there is a relational path associated with the clause $C_k$ from $X_i$ to $Y_i$.
Furthermore, let  
$\textup{neg}_k$ be a number associated  with the number of ``nonedges'' $(X', Y')$ from $(V \times V) \setminus E_r$ for which $C_k(X',Y') = 1$.
We calculate $\textup{neg}_k$ for the $k$th rule as follows.
We consider the tail node $t$ and head node $h$ for each edge in $E_r$. We compute the set of nodes $S$ that can be reached by a path induced by the $k$th rule starting at the tail. If there is no edge from $t$ to a node $v$ in $S$ labeled by $r$, we say that $v$ is an invalid end-point. Let $\textup{right}_k$ be the set of such invalid points. We similarly calculate the set $\textup{left}_k$ of invalid start-points based on paths ending at $h$ induced by the $k$th rule. The total number of invalid start and end points for all tail and head nodes associated with edges in $E_r$ is $\textup{neg}_k = |\textup{right}_k| + |\textup{left}_k|$. For a query of the form $(t, r, ?)$ where $t$ is a tail node of an edge in $E_r$, the scoring function defined by the $k$th rule alone gives a positive and equal score to all nodes in $\textup{right}_k$.

Our model for rule-learning is given below.
\begin{eqnarray}
\mbox{(LPR)}  \nonumber\\
 z_{min} = & \min &  \sum_{i=1}^m \eta_i + \tau\sum_{k \in \K} \textup{neg}_kw_k \label{obj1} \\
  & s.t. &\sum_{k \in \K} a_{ik} w_k + \eta_i \geq 1 \mbox{ for all } i \in E_r \label{LPR_const1}\\
 & & \sum_{k \in \K} (1 + |C_k|) w_k \leq \kappa \label{LPR_const2}\\
    & & w_k \in [0,1] \mbox{ for all } k \in \K\label{LPR_bounds}\\
    & & \eta_i \geq 0 \mbox{ for all } i \in E_r.
\end{eqnarray}    
The variable $w_k$ is restricted to lie in $[0,1]$ and is positive if and only if clause $k \in \K$ is a part of the scoring function (\ref{pred1}). The parameter $\kappa$ is an upper bound on the {\em complexity} of the scoring function (defined as the number of clauses plus the number of relations across all clauses).
$\eta_i$ is a penalty variable which is positive if the scoring function defined by positive $w_k$s gives a value less than 1 to the $i$th edge in $E_r$. Therefore, the $\sum_{i=1}^m \eta_i$ portion of the objective function attempts to maximize $\sum_{i=1}^m \min\{f_r(X_i, Y_i), 1\}$, i.e., it attempts to approximately maximize the number of facts in $E_r$ that are given a ``high-score" of 1 by $f_r$. 
In addition, we have the parameter $\tau > 0$ which represents a tradeoff between how well our weighted combination of rules performs on the known facts (gives positive scores), and how poorly it performs on some ``missing" facts. We make this precise shortly. 
Maximizing the MRR is a standard objective for KGC and thus the objective function of LPR is only an approximation; see the next Theorem. We still obtain high-quality prediction rules using LPR.

\begin{thm}\label{thm1}
Let IPR be the integer programming problem created from LPR by replacing equation (\ref{LPR_bounds}) by $w_k \in \{0,1\}$ for all $k \in \K$, and letting $\tau = 0$. Given an optimal solution with objective function value $\gamma$, one can construct a scoring function such that $1-\gamma/m$ is a lower bound on the MRR of the scoring function calculated by the optimistic ranking method, when applied to the training set triples.
\end{thm}
The theorem above justifies choosing IPR as an optimization formulation to find good rule sets for a relation, assuming MRR calculation via optimistic ranking.
When using random break ranking, it is essential to perform negative sampling and penalize rules that create paths when there are no edges in order to produce good quality results. This is why we use $\tau> 0$ in LPR. 
We will now give an interpretation of $\sum_k  \textup{neg}_k w_k$. To compute the MRR of the scoring function $f_r$ in (\ref{pred1}) applied to the training set, for each edge $(t,r,h) \in E_r$ we need to compute the rank of the answer $h$ to the query $(t,r,?)$ -- by comparing $f_r(t,v)$ with $f_r(t,h)$ for all nodes $v$ in $\mathcal{G}$ -- and the rank of answer $t$ to the query $(?,r,h)$ -- by comparing $f_r(v,h)$ with $f_r(t,h)$ for all nodes $v$. But $\sum_k \textup{neg}_k w_k$ is exactly the sum of scores given by $f_r$ to all nodes in $\textup{right}_k$ and $\textup{left}_k$ and therefore we have the following remark.
\begin{remark}
Let $(t, r, h)$ be an edge in $E_r$, and let $U(?,r,h)$ and $U(t,r,?)$ be the set of invalid answers for $(?, r, h)$ and $(t, r, ?)$, respectively. Then
\begin{eqnarray} \sum_{(t,r,h) \in E_r} \left(\sum_{v \in U(?,r,h)} f_r(v,h) + \sum_{v \in U(t,r,?)} f_r(t, v) \right) \nonumber\\
 = \sum_{k \in \mathcal{K}} \textup{neg}_k w_k. \nonumber
\end{eqnarray}
\end{remark}
In other words, rather than keeping individual scores of the form $f_r(v,h)$ and $f_r(t,v)$ for missing facts $(v,r,h)$ or $(t,r,v)$ small, we minimize the sum of these scores in LPR.

It is impractical to solve LPR given the exponentially many variables $w_k$, except when $n$ and $L$ are both small. 
An effective way to solve such large LPs is to use column generation 
where only a
small subset of all possible $w_k$ variables is generated explicitly and the optimality of the LP
is guaranteed by iteratively solving a {\em pricing} problem. We do not attempt to solve LPR to optimality.
We start with an initial set of candidate rules $\K_0 \subset \K$ (and implicitly set all rule variables from $\K \setminus \K_0$ to 0). Let LPR$_0$ be the associated LP. If the solution of LPR$_0$ is not satisfactory, we  dynamically augment the set of candidate rules to create sets $\K_i$ such that $\K_0 \subset \K_1 \subset \cdots \subset \K$. If LPR$_i$ is the LP associated with $\K_i$ with optimal solution value $z^i_{min}$, then it is clear that a solution of LPR$_i$ yields a solution of LPR$_{i+1}$ by setting the extra variables in LPR$_{i+1}$ to zero, and therefore $z^{i+1}_{min} \leq z^i_{min}$. We attempt to have $z^{i+1}_{min} < z^i_{min}$
by taking the dual solution associated with an optimal solution of LP$_{i}$, and then trying to find a {\em negative reduced cost} rule, which we discuss shortly.

\begin{table}[h]
\centering
\resizebox{\linewidth}{!}{
\begin{tabular}{lrrrrr}  
\toprule
Datasets & \# Relations & \# Entities & \# Train & \# Test & \# Valid \\
\midrule
Kinship	 & 25	    & 104    & 8544  & 1074 & 1068  \\[5pt]
UMLS     & 46       & 135    & 5216  & 	661 & 652   \\[5pt]
WN18RR	 & 11       & 40943  & 86835 & 3134 & 3034  \\[5pt]
FB15k-237& 237      & 14541  & 272115& 20466& 17535 \\[5pt]
YAGO3-10 & 37       & 123182 &1079040& 5000 & 5000  \\[2pt]
\bottomrule 
\end{tabular}
}
\caption{Sizes of datasets.}
\label{table3}
\end{table}

\begin{table*}[h]
\centering
\small
\begin{tabular}{llrrrrrrrr}
Problem & metric & ComplEx-N3 & TuckER & $\dagger$ConvE & AnyBURL & NeuralLP & DRUM & RNNLogic & LPRules \\
\hline
Kinship  
 & MRR  & 0.889 & 0.891 & 0.83 & 0.626 & 0.652 & 0.566 & {\it 0.687} & {\bf 0.746}  \\[5pt]
 & H@1  & 0.824 & 0.826 & 0.74 & 0.503 & 0.52  & 0.404 & {\it 0.566}  & {\bf 0.639}  \\[5pt]
 & H@10 & 0.986 & 0.987 & 0.98 & 0.901 & 0.925 & 0.91  & {\it 0.929}  & {\bf 0.959}  \\[5pt]
\hline
UMLS
 & MRR  & 0.962 & 0.914 & 0.94 & {\bf 0.940} & 0.75  & 0.845 & 0.748 & {\it 0.869}  \\[5pt]
 & H@1  & 0.934 & 0.837 & 0.92 & {\bf 0.916} & 0.601 & 0.722 & 0.618  & {\it 0.812}  \\[5pt]
 & H@10 & 0.996 & 0.997 & 0.99 & {\it 0.985} & 0.958 & {\bf 0.991} & 0.928  & 0.97   \\[5pt]
\hline
FB15K-237
 & MRR  & 0.362 & 0.353 & 0.325 & 0.226 & 0.222 & 0.225 & $\sharp${\bf 0.288} & {\it 0.255} \\[5pt]
 & H@1  & 0.259 & 0.259 & 0.237 & 0.166 & 0.16  & 0.16  & {\bf 0.208}  & {\it 0.17}  \\[5pt]
 & H@10 & 0.555 & 0.538 & 0.501 & 0.387 & 0.34 & 0.355 & {\bf 0.445}  & {\it 0.402} \\[5pt]
\hline
WN18RR
 & MRR  & 0.469 & 0.464 & 0.43  & {\it 0.454} & 0.381 & 0.381 & 0.451 & {\bf 0.459} \\[5pt]
 & H@1  & 0.434 & 0.436 & 0.4   & {\bf 0.423} & 0.367 & 0.367 & 0.415 & {\it 0.422} \\[5pt]
 & H@10 & 0.545 & 0.517 & 0.52  & {\it 0.527} & 0.409 & 0.41  & 0.524  & {\bf 0.532} \\[5pt]
\hline
YAGO3-10
 & MRR  & 0.574 & 0.265 & 0.44  & {\bf 0.449} &       &       &       &  {\bf 0.449} \\[5pt]
 & H@1  & 0.499 & 0.184 & 0.35  & {\bf 0.381} &       &       &       &  {\it 0.367} \\[5pt]
 & H@10 & 0.705 & 0.426 & 0.62  & {\it 0.598} &       &       &       &  {\bf 0.684} \\[5pt]
\hline
\end{tabular}
\caption{Comparison of results on standard datasets. The results for NeuralLP, DRUM, RNNLogic, LPRules use the random break metric. $\dagger$ ConvE results are from \citet{dettmers2018ConvE}. $\sharp$ We could not run RNNLogic on FB15k-237, and report numbers from \citet{RNNLogic}. We run the {\em light} version of AnyBURL which yields entity-independent rules only. The best and second best values in any row for rule-based methods are given in bold and italics, respectively.}
\label{table6}
\end{table*}

\noindent {\bf Setting up the initial LP.} 
To set up $\K_0$ and LP$_0$, 
we develop two heuristics. In Rule Heuristic 1, we generate rules of lengths one and two for a relation $r$. We create a one-relation rule from $r' \in \mathcal{R} \setminus \{r\}$ if it labels a large number of edges from tail nodes to head nodes of edges in $E_r$. We essentially enumerate the length-two rules $r_1(X,Y) \wedge r_2(Y,Z)$ and keep those that frequently create paths from the tail nodes to head nodes of edges in $E_r$.
In Rule Heuristic 2, we take each edge $(X,Y)$ in $E_r$ and find a shortest path from $X$ to $Y$ contained in the edge set $E \setminus \{(X,Y)\}$ where the path length is bounded by a pre-determined maximum length.  We then use the sequence of relations associated with the shortest path to generate a rule. We also use a path of length at least one more than the shortest path.

\noindent {\bf Adding new rules.} Each $\K_i$ for $i>0$ is constructed by adding new rules to $\K_{i-1}$. We use a modified version of Heuristic 2 to generate the additional rules.
Let $\delta_i \geq 0$ for all $i \in E_r$ be dual variables corresponding to constraints (\ref{LPR_const1}).
Let $\lambda \leq 0$ be the dual variable associated with the constraint (\ref{LPR_const2}).
Given a variable $w_k$ which is zero in a solution of LPR$_i$ and dual solution values $\bar\delta$ and $\bar \lambda$ associated with the optimal solution of LPR$_{i-1}$, the reduced cost $\textup{red}_k$ for $w_k$ is
\begin{equation*}
\textup{red}_k = \tau \: \textup{neg}_k - \sum_{i \in E_r} a_{ik} \bar \delta_i - (1 + |C_k|) \bar \lambda
\end{equation*}
If $\textup{red}_k < 0$, then increasing $w_k$ from zero may reduce the LP solution value.
In our heuristic, we sort the dual values $\bar\delta_j$ in decreasing order, and then go through the associated indices $j$, and create rules $k$ such that $a_{jk} = 1$ via a shortest path calculation.
That is, we
take the corresponding edge $(X,Y)$ in $E_r$, find the shortest path between $X$ and $Y$ and generate a new rule with the sequence of relations in that path. We limit the number of rules generated so that $|\K_i| - |\K_{i-1}| \leq 10$.
 We do not add new rules to $\K_{i-1}$ if the reduced cost of the new rule is nonnegative.
 Our overall method is described in Algorithm~\ref{alg:algorithm}.

\begin{algorithm}[tb]
\caption{LPRules}
\label{alg:algorithm}
\textbf{Input}: Train, test, \& validation datasets.  \\
\textbf{Parameters}: Sets of $\tau$ values \& complexity bounds $\kappa$. \\
\textbf{Output}: Rules \& evaluation metrics.
\begin{algorithmic}[1] 
\FOR{each relation $r$ in train}
\STATE Call Rule Heur. 1 \& 2 to generate initial rule set $\K_0$.
\STATE Set up LPR$_0$ from $\K_0$ using min $\tau$ \& min $\kappa$ \& solve.
\STATE Set $\text{bestMRR}=0$, $\text{best}\kappa=\text{best}\tau=-\infty$.
\FOR{$i \gets 1$ to MaxIter}
\STATE Generate new rules with modified Rule Heur. 2.
\STATE Add new rules with reduced cost $< 0$ to $\K_{i-1}$ to form $\K_i$, set up LPR$_i$ from $\K_i$ \& solve.
\ENDFOR
\FOR{each $\tau$}
\FOR{each $\kappa$}
\STATE Set up LPR$_i$ from $\K_i$ using current value of $\tau$ \& $\kappa$ \& solve.
\STATE Use soln \& compute MRR on validation facts.
\IF{MRR $>$ bestMRR}
\STATE bestMRR = MRR.
\STATE best$\tau$ = current $\tau$, best$\kappa$ = current $\kappa$.
\ENDIF
\ENDFOR
\ENDFOR
\STATE Set up LPR$_i$ from $\K_i$ using best$\tau$, best$\kappa$ \& solve.
\STATE Output rules with corresponding weights in LPR$_i$.
\ENDFOR
\STATE Compute evaluation metrics on test set.
\end{algorithmic}
\end{algorithm}

\section{Experiments}

We conduct KGC experiments with 5 datasets: Kinship \citep{denham}, UMLS \citep{mccray}, FB15k-237 \citep{toutanova}, WN18RR \citep{dettmers2018ConvE}, and YAGO3-10 \citep{Mahdisoltani2015YAGO3AK}.
The partition of FB15k-237, WN18RR, and YAGO3-10 into training, testing, and validation data sets is standard. We use the partition for UMLS and Kinship in \citet{dettmers2018ConvE}.
In Table \ref{table3}, we give the number of entities and relations in each dataset, and facts in each partition. 

\subsection{Experimental setup}\label{expsetup}

We denote the reverse relation for $r \in \mathcal{R}$ by $r^{-1}$. For each fact $(t,r,h)$ in the training set, we implicitly introduce the fact $(h, r^{-1}, t)$.
For each {\em original} relation $r$ in the training set, we create a scoring function $f_r(X,Y)$  of the form in (\ref{pred1}).
For each test set fact $(t,r,h)$ we create two queries $(t,r,?)$ and $(?, r, h)$.
We compute $f_r(t,e)$ and $f_r(e,h)$ for each entity $e$ in $\G$.
We use the filtered ranking method in \citet{bordes} to calculate a rank for the correct answer $(t,r,h)$, which is then used to compute MRR and Hits@$k$ (for $k=1,10$) across all test facts.
Ties between equal scores are broken using the random break method.

We compare our results with the rule-based methods AnyBURL, NeuralLP, DRUM, and RNNLogic. 
We use default settings for  NeuralLP and DRUM and the settings proposed by the authors of RNNLogic. We modify RNNLogic to implement the random break method to break ties.
AnyBURL uses a lexicographic method to break ties; we run it for 100 seconds with the {\em light} settings which cause AnyBURL to only generate entity-independent rules.
We give results for the popular embedding-based methods ConvE \citep{dettmers2018ConvE}, 
ComplEx-N3 \citep{pmlr-v80-lacroix18a}, and TuckER \citep{balazevic2019tucker}, 
mainly to show the maximum achievable MRR on our datasets. We ran ComplEx-N3 and TuckER on our machines with the best published hyperparameters (if available). 
ConvE results are taken from \citet{dettmers2018ConvE}. 

We ran two variants of our code which we call ``LPRules"\footnote{https://github.com/IBM/LPRules} (see Algorithm~\ref{alg:algorithm}). In the first variant, we create LPR$_0$ by generating rules using Rule Heuristics 1 and 2, and then solve LPR$_0$ to obtain rules (thus MaxIter = 0). In the second variant (used only for YAGO3-10) we create LPR$_0$ with an empty set of rules and then perform 15 iterations consisting of generating up to 10 rules using the modified version of Rule Heuristic 2 followed by solving the new LP. In other words, we create and solve LPR$_i$ for $i=0, \ldots 15$. To compute $\textup{neg}_k$, we sample 2\% of the edges from $E_r$, and compute the number of invalid paths that start at the tails of these edges, and end at the heads of these edges. 
We search for the best values of $\tau$ and $\kappa$ for each relation, where $\tau$ belongs to an input set of values.  
We dynamically let $\bar \kappa$ equal the length of the longest rule generated plus one. We then perform 20 iterations where, at the $i$th iteration, we set $\kappa$ to $i\bar\kappa$. We use the validation data set to select those $\tau$ and $\kappa$ that yield the best MRR. We set the maximum rule length to 6 for WN18RR, and 3 for YAGO3-10, and 4 for the other datasets. Thus $\kappa \leq 100$ except for WN18RR.  

\subsection{Results}

We run the rule-based codes on a 60 core machine with 128 GBytes of RAM, and four 2.8 Intel Xeon E7-4890 v2 processors, each with 15 cores. 
In our code, we execute rule generation for each relation on a different thread, and solve LPs with CPLEX \citep{cplex}.
If one relation has many more facts than the others, as in YAGO3-10, then our code essentially uses one thread.

In Table \ref{table6}, we give values for different metrics obtained with the listed codes, first for embedding methods, then for rule-based methods (if available), and then for our code. We could not run RNNLogic on FB15k-237 and use the published result.
The best embedding method is better across all metrics than rule-based methods.
Our method is the best rule-based code across all metrics on Kinship, obtains better MRR and Hits@10 values on WN18RR and YAGO3-10, and second best values for FB15K-237 (and MRR and Hits@1 numbers for UMLS). This is notable as we use very simple rule generation heuristics, generate relatively compact rules, and take 
 significantly less computing time than DRUM, NeuralLP, and RNNLogic (see Table~\ref{table9}).
 These three codes are unable to produce results for YAGO3-10 within a reasonable time frame.
 For YAGO3-10, our column generation approach, where we  generate a small number of rules, and then use the dual values to focus on ``uncovered" facts (not implied by previous rules) is essential.

\begin{table}[ht]
\centering
\small
\resizebox{\linewidth}{!}{
\begin{tabular}{llcccc}  
\toprule
Problem & metric & AnyBURL & NeuralLP & RNNLogic & LPRules \\
\midrule
Kinship &  rules  & 6653.1  & 10.4  & 200.0  & 21.0  \\[1pt]
UMLS    &  rules  & 1837.6  & 15.1  & 100.0  &  4.2  \\[1pt]
FB15k-237  & rules &  79.9  &  8.1  &     -   & 14.2  \\[1pt]
WN18RR     & rules &   47.3  & 14.3  & 200.0  & 15.6  \\[1pt]
YAGO3-10   & rules &   63.0  & -   &        &  7.8  \\[1pt]
\midrule
Kinship & time      & 1.7     & 1.6   & 108.8  & 0.5  \\[1pt]
UMLS & time      & 1.9     & 1.1   & 133.4  &  0.2  \\[1pt]
FB15k-237 & time    &  3.9   &  14565.9    &    -    & 234.5  \\[1pt]
WN18RR & time    &   1.8   & 399.9 & 104.0  & 11.0  \\[1pt]
YAGO3-10 & time    &   34.3  & -   &        &  1648.4  \\[1pt]
\bottomrule 
\end{tabular}
}
\caption{Average number of rules selected per relation and total wall clock running time in minutes when running in parallel on a 60 core machine.}
\label{table9}
\end{table}

In Table~\ref{table9}, we compare the average number of rules per relation (we could not extract rules from DRUM) in the final solution and the running times (DRUM has comparable running time to NeuralLP). Our code always obtains Pareto optimal solutions (when measured on MRR and average number of rules). For WN18RR and YAGO3-10, we obtain the best MRR values with few rules: we get excellent results for YAGO3-10 with just 7.8 rules on average. 

\begin{figure*}[htbp]
\centering
  \includegraphics[width=.49\linewidth]{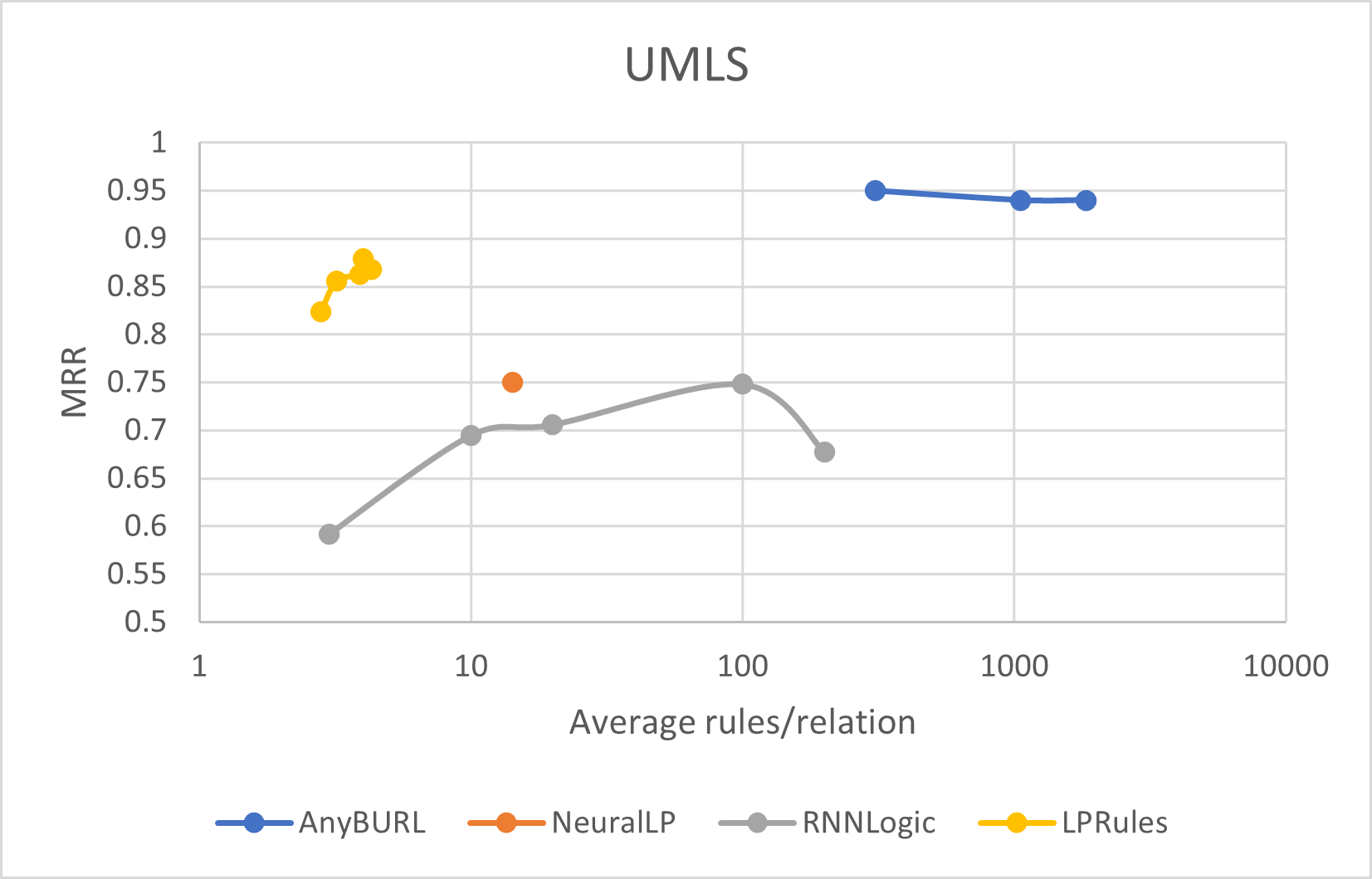}
  \includegraphics[width=.49\linewidth]{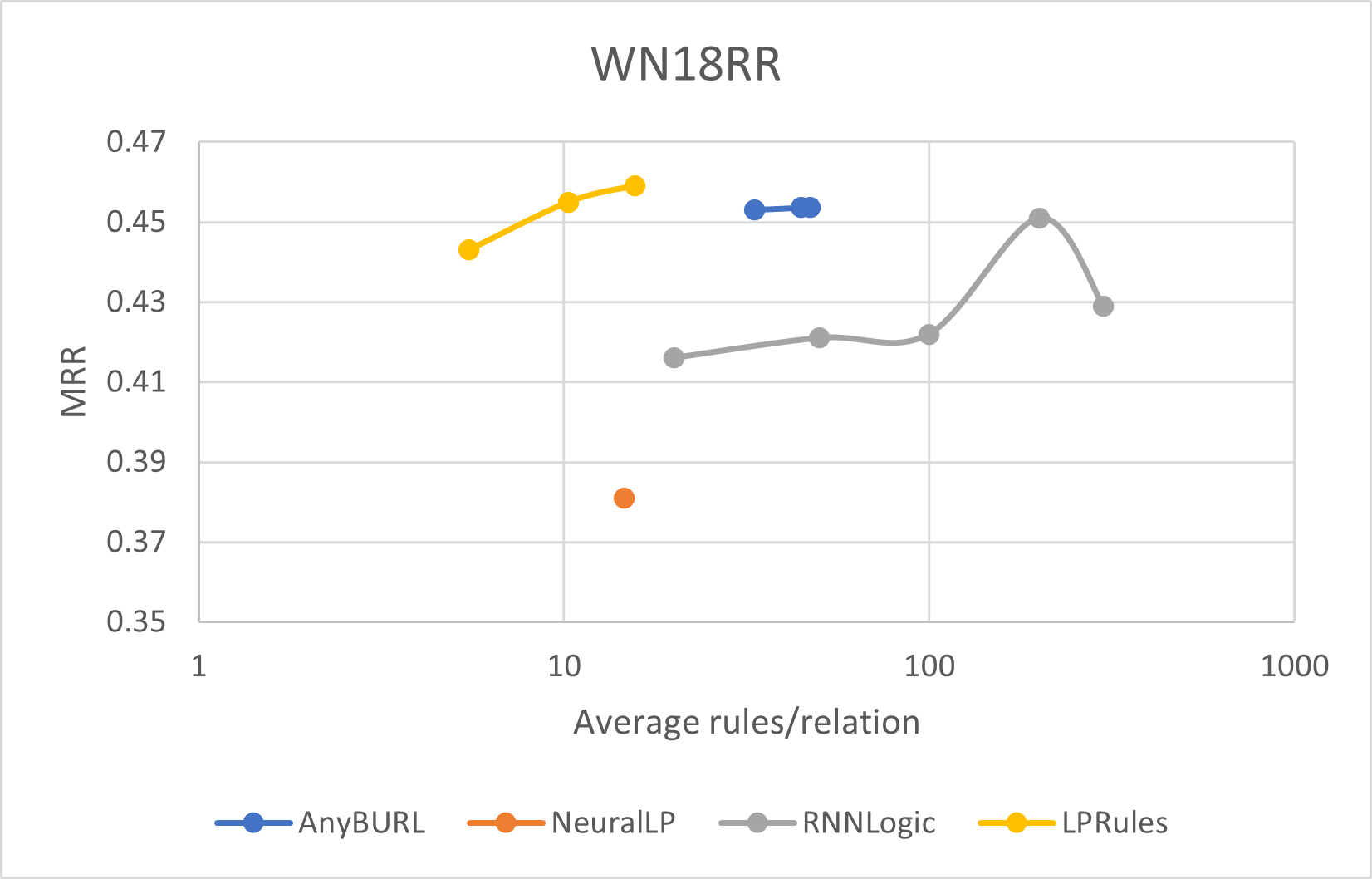}
\caption{Change in MRR with change in average number of rules per relation for (a) UMLS and (b) WN18RR}
\label{fig:graph_vary_rules} 
\end{figure*}

In Figure~\ref{fig:graph_vary_rules}, we show how MRR varies with average number of rules in the solution.
RNNLogic chooses top-$K$ rules for testing ($K$ is an input parameter), and we run it with different values of $K$. For AnyBURL, we take the rules generated at 10, 50, and 100 seconds. We are unable to control the output number of rules in NeuralLP. Figure~\ref{fig:graph_vary_rules} (b) is especially striking and demonstrates a much better MRR to number of rules tradeoff than RNNLogic on WN18RR.

\begin{figure*}[ht]
\centering
  \includegraphics[width=.49\linewidth]{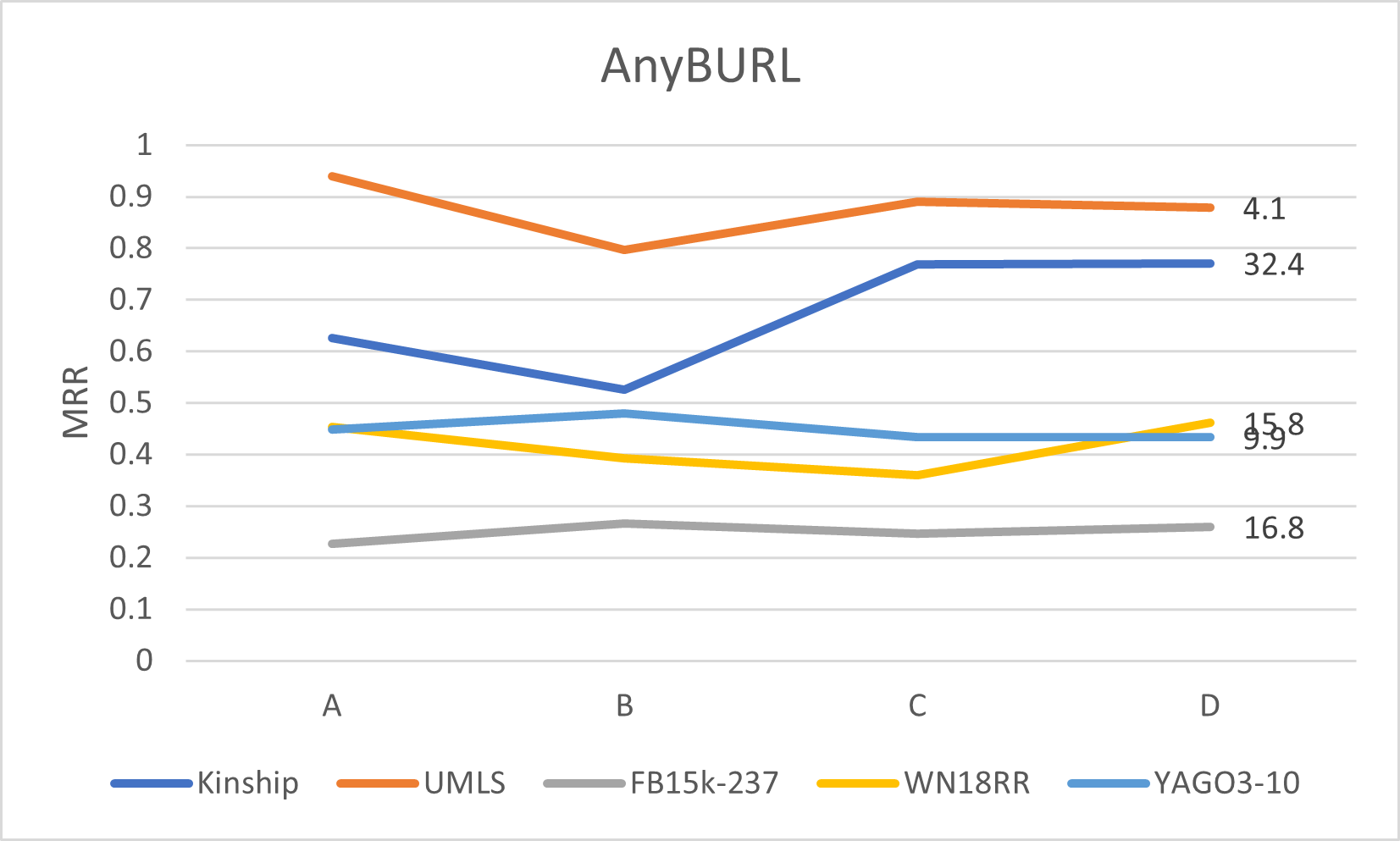}
  \includegraphics[width=.49\linewidth]{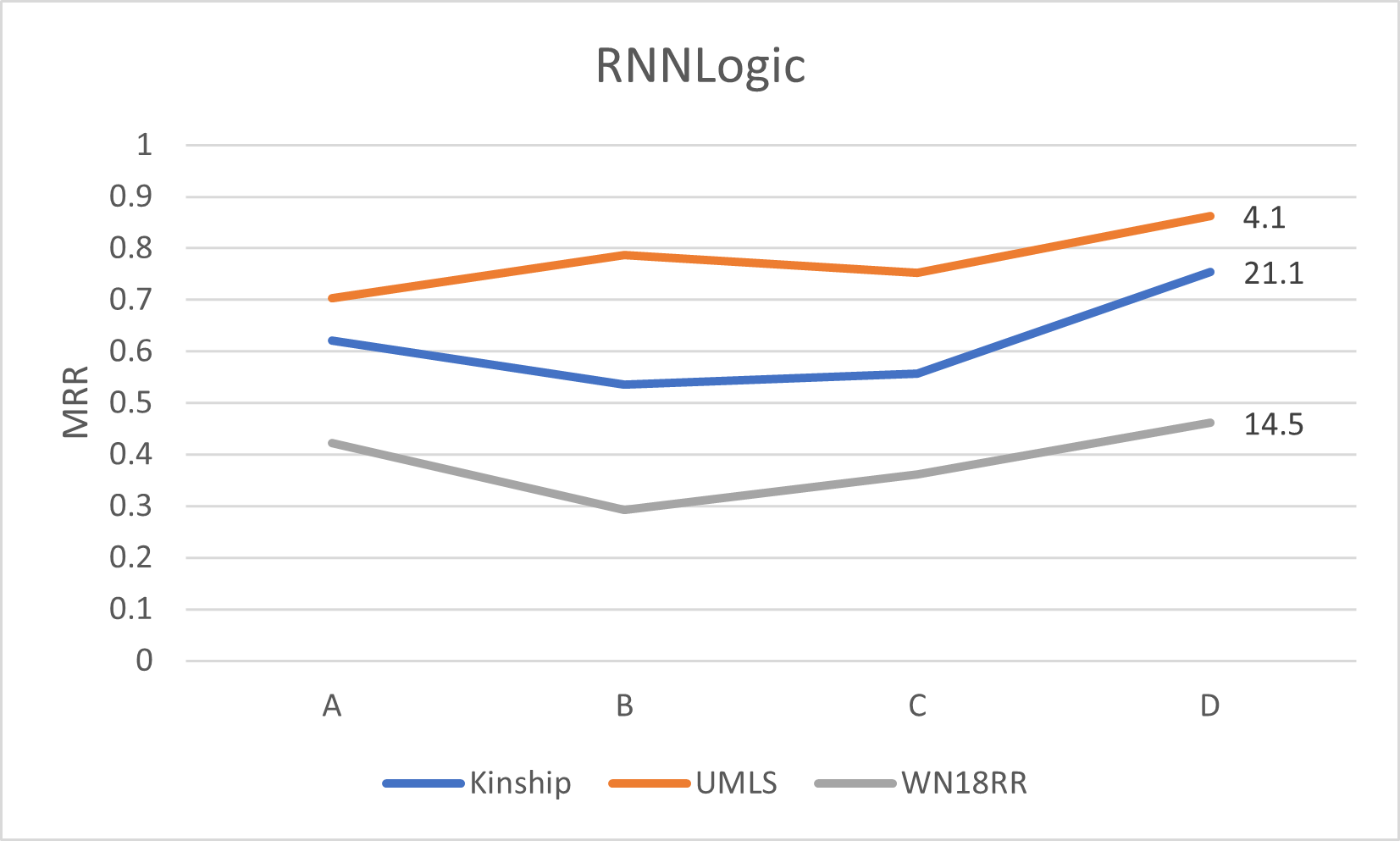}
\caption{MRR values in different scenarios when using rules generated by (a) AnyBURL and (b) RNNLogic. MRR is shown on the $y$-axis. The number to the right of each curve is the average number of rules per relation in the final solution in Scenario D.}
\label{fig:graph_anyburl_rnn} 
\end{figure*}

Our LP formulation can be initiated with {\em any input candidate set} of rules.
In Figure~\ref{fig:graph_anyburl_rnn} (a) and (b), we show the effect of combining rules generated by AnyBURL and RNNLogic with our rules. The $y$-axis corresponds to MRR values, and the letters A,B,C,D on the $x$-axis stand for four scenarios. In Scenario A, we run another rule-based code. In Scenario B, we take as input the rules and rule-weights from Scenario A to build our scoring function. In Scenario C, we give these rules to our LP formulation, and recalculate weights, while limiting the complexity of the final solution. Finally, in Scenario D, we give the rules in Scenario C along with our heuristically generated rules to our LP formulation. 
Our scoring function is neither better nor worse than those of AnyBURL and RNNLogic (or NeuralLP). Using the same rules and rule weights generated by AnyBURL (in Scenario A), our scoring function produces (in Scenario B) better MRR in two cases and worse in three cases. Just using our scoring function instead of AnyBURL's, we get an MRR of {\bf 0.267} (instead of 0.226) for FB15K-237, and an MRR of {\bf 0.480} (instead of 0.449) for YAGO3-10; these values are better than those obtained by either AnyBURL or LPRules. 
We get better MRR values in Scenario C than AnyBURL for Kinship and FB15K-237 with much more compact solutions. That is we choose a subset of the AnyBURL rules, give different weights, and yet get a better solution. It is hard to see a trend going from Scenario B to C. We conclude that our LP approach can combine rules generated by other methods with our rules, and get similar or larger MRR values even while choosing few rules.

\begin{table}[h]
\centering
\small
\resizebox{\linewidth}{!}{
\begin{tabular}{lcccccccc}  
\toprule
 & & \multicolumn{3}{c}{GraIL}  & &\multicolumn{3}{c}{LPRules}  \\
\cmidrule{3-5} \cmidrule{7-9}
 Problem & Ver. & MRR & H@1 & H@10 && MRR & H@1 & H@10\\
\midrule
WN18RR     & v1  & 0.556 & 0.434 & 0.769 && 0.682 & 0.636 & 0.745 \\[5pt]
           & v2  & 0.587 & 0.497 & 0.734 && 0.646 & 0.609 & 0.700 \\[5pt]
           & v3  & 0.303 & 0.254 & 0.391 && 0.398 & 0.359 & 0.460 \\[5pt]
           & v4  & 0.533 & 0.456 & 0.671 && 0.609 & 0.577 & 0.668 \\[5pt]
\midrule
FB15k-237  & v1  & 0.238 & 0.173 & 0.339 && 0.313 & 0.271 & 0.380 \\[5pt]
           & v2  & 0.203 & 0.129 & 0.328 && 0.396 & 0.315 & 0.539 \\[5pt]
\bottomrule 
\end{tabular}
}
\caption{Results on inductive datasets from \cite{teru}. We were not able to obtain results for GraIL on FB15k-237\_v3 and FB15k-237\_v4.}
\label{table_s9}
\end{table}

Entity-independent rules have a strong inductive bias, especially when very few rules are generated per relation. We compare our code on an inductive benchmark 
dataset with GraIL \cite{grail}, which learns the entity-independent subgraph structure around the edges associated with each relation. Relational paths corresponding to the chain-like Horn rules we use  are special cases of the subgraphs learnt by GraIL.  A low-quality solution of the much harder learning problem in GRAIL could lead to worse predictions than a high-quality solution of our simpler learning problem. We indeed obtain better results than GraIL on its inductive benchmarks, see Table~\ref{table_s9}. Each dataset is split into a KG for training and a KG for testing that has a subset of the training KG relations, but no common entities.  All GraiIL results were obtained with the parameter ``Negative Sampling Mode'' set to ``all''.

\section{Conclusion}

Most rule-based methods do not focus on compact rule sets. Our relatively simple LP based method for selecting weighted logical rules can return state-of-the-art results for a number of standard KG datasets even with compact (and potentially more interpretable) rule sets. Further it has better scaling than a number of existing neuro-symbolic methods. Finally, we demonstrate that our method yields better results in an inductive setting than a recent solver that performs subgraph reasoning.
Our work can be improved further in several areas such as accuracy and scaling with KG size. To improve scaling, one can sample facts when dealing with large KGs in order to obtain LPs of  manageable size and also process different goups of facts on different machines. As demonstrated in Figure 2, accuracy can be improved by generating additional rules using different algorithms. Handling ontologies, more complex queries, and more general rules would be other natural extensions of our work.

\bibliography{references}

\begin{thebibliography}{43}
\providecommand{\natexlab}[1]{#1}

\bibitem[{Bala\v{z}evi\'c, Allen, and Hospedales(2019)}]{balazevic2019tucker}
Bala\v{z}evi\'c, I.; Allen, C.; and Hospedales, T.~M. 2019.
\newblock TuckER: Tensor Factorization for Knowledge Graph Completion.
\newblock In \emph{Empirical Methods in Natural Language Processing 2019}.

\bibitem[{Bordes et~al.(2013)Bordes, Usunier, Garcia-Duran, Weston, and
  Yakhnenko}]{bordes}
Bordes, A.; Usunier, N.; Garcia-Duran, A.; Weston, J.; and Yakhnenko, O. 2013.
\newblock Translating Embeddings for Modeling Multi-relational Data.
\newblock In \emph{Neurips 2013}.

\bibitem[{Cropper and Morel(2020)}]{learn-fail}
Cropper, A.; and Morel, R. 2020.
\newblock Learning programs by learning from failures.
\newblock arXiv:2005.02259.

\bibitem[{Cropper and Muggleton(2016)}]{metagol-ai}
Cropper, A.; and Muggleton, S.~H. 2016.
\newblock Learning Higher-Order Logic Programs Through Abstraction and
  Invention.
\newblock In \emph{IJCAI 2016}.

\bibitem[{Das et~al.(2018)Das, Dhuliawala, Zaheer, Vilnis, Durugkar,
  Krishnamurthy, Smola, and McCallum}]{minerva}
Das, R.; Dhuliawala, S.; Zaheer, M.; Vilnis, L.; Durugkar, I.; Krishnamurthy,
  A.; Smola, A.; and McCallum, A. 2018.
\newblock Go for a walk and arrive at the answer: Reasoning over paths in
  knowledge bases using reinforcement learning.
\newblock In \emph{ICLR 2018}.

\bibitem[{Dash, G\"unl\"uk, and Wei(2018)}]{dash2018boolean}
Dash, S.; G\"unl\"uk, O.; and Wei, D. 2018.
\newblock Boolean decision rules via column generation.
\newblock In \emph{Advances in Neural Information Processing Systems},
  4655--4665.

\bibitem[{Demiriz, Bennett, and Shawe-Taylor(2002)}]{dbs}
Demiriz, A.; Bennett, K.~P.; and Shawe-Taylor, J. 2002.
\newblock Linear programming boosting via column generation.
\newblock \emph{Machine Learning}, 46: 225--254.

\bibitem[{Denham(1973)}]{denham}
Denham, W. 1973.
\newblock \emph{The detection of patterns in Alyawarra nonverbal behavior}.
\newblock Ph.D. thesis, University of Washington.

\bibitem[{Dettmers et~al.(2018)Dettmers, Pasquale, Pontus, and
  Riedel}]{dettmers2018ConvE}
Dettmers, T.; Pasquale, M.; Pontus, S.; and Riedel, S. 2018.
\newblock Convolutional 2D Knowledge Graph Embeddings.
\newblock In \emph{AAAI}.

\bibitem[{Eckstein and Goldberg(2012)}]{eg2}
Eckstein, J.; and Goldberg, N. 2012.
\newblock An Improved Branch-and-Bound Method for Maximum Monomial Agreement.
\newblock \emph{INFORMS Journal on Computing}, 24(2): 328--341.

\bibitem[{Eckstein, Kagawa, and Goldberg(2019)}]{eg3}
Eckstein, J.; Kagawa, A.; and Goldberg, N. 2019.
\newblock REPR: Rule-Enhanced Penalized Regression.
\newblock \emph{INFORMS Journal on Optimization}, 1(2): 143--163.

\bibitem[{Evans and Grefenstette(2018)}]{evans}
Evans, R.; and Grefenstette, E. 2018.
\newblock Learning explanatory rules from noisy data.
\newblock \emph{Journal of Artificial Intelligence Research}, 61: 1--64.

\bibitem[{IBM(2019)}]{cplex}
IBM. 2019.
\newblock {IBM} {CPLEX} Optimizer, version 12.10.

\bibitem[{Ji et~al.(2021)Ji, Pan, Cambria, Marttinen, and Yu}]{kg-survey}
Ji, S.; Pan, S.; Cambria, E.; Marttinen, P.; and Yu, P.~S. 2021.
\newblock A Survey on Knowledge Graphs: Representation, Acquisition, and
  Applications.
\newblock \emph{IEEE Transactions on Neural Networks and Learning Systems},
  1--21.

\bibitem[{Kok and Domingos(2005)}]{kok}
Kok, S.; and Domingos, P. 2005.
\newblock Learning the structure of Markov Logic Networks.
\newblock In \emph{ICML 2005}.

\bibitem[{Kovalerchuk, Ahmad, and Teredesai(2021)}]{iai}
Kovalerchuk, B.; Ahmad, M.; and Teredesai, A. 2021.
\newblock \emph{Survey of Explainable Machine Learning with Visual and Granular
  Methods Beyond Quasi-Explanations}, 217--267.
\newblock Cham.: Springer.

\bibitem[{Lacroix, Usunier, and Obozinski(2018)}]{pmlr-v80-lacroix18a}
Lacroix, T.; Usunier, N.; and Obozinski, G. 2018.
\newblock Canonical Tensor Decomposition for Knowledge Base Completion.
\newblock In \emph{ICML 2018}.

\bibitem[{Lao and Cohen(2010)}]{lao-cohen}
Lao, N.; and Cohen, W.~W. 2010.
\newblock Relational retrieval using a combination of path-constrained random
  walks.
\newblock \emph{Machine Learning}, 81: 53--67.

\bibitem[{Lin, Socher, and Xiong(2018)}]{multihopkg}
Lin, X.~V.; Socher, R.; and Xiong, C. 2018.
\newblock Multi-hop knowledge graph reasoning with reward shaping.
\newblock In \emph{Empirical Methods in Natural Language Processing 2018}.

\bibitem[{Mahdisoltani, Biega, and Suchanek(2015)}]{Mahdisoltani2015YAGO3AK}
Mahdisoltani, F.; Biega, J.; and Suchanek, F.~M. 2015.
\newblock YAGO3: A Knowledge Base from Multilingual Wikipedias.
\newblock In \emph{CIDR}.

\bibitem[{McCray(2003)}]{mccray}
McCray, A.~T. 2003.
\newblock An upper level ontology for the biomedical domain.
\newblock \emph{Comparative and Functional Genomics}, 4: 80--84.

\bibitem[{Mehrabi et~al.(2022)Mehrabi, Morstatter, Saxena, Lerman, and
  Galstyan}]{bias}
Mehrabi, N.; Morstatter, F.; Saxena, N.; Lerman, K.; and Galstyan, A. 2022.
\newblock A Survey on Bias and Fairness in Machine Learning.
\newblock \emph{ACM Computing Surveys}, 54: 1--35.

\bibitem[{Meilicke et~al.(2019)Meilicke, Chekol, Ruffinelli, and
  Stuckenschmidt}]{anyburl}
Meilicke, C.; Chekol, M.~W.; Ruffinelli, D.; and Stuckenschmidt, H. 2019.
\newblock Anytime bottom-up rule learning for knowledge graph completion.
\newblock \emph{IJCAI 2019}.

\bibitem[{Michie(1988)}]{michie}
Michie, D. 1988.
\newblock Machine learning in the next five years.
\newblock Proceedings of the Third European Working Session on Learning,
  107–--122. Pitman.

\bibitem[{Muggleton et~al.(2018)Muggleton, Schmid, Zeller, Tamaddoni-Nezhad,
  and Besold}]{mug2}
Muggleton, S.~H.; Schmid, U.; Zeller, C.; Tamaddoni-Nezhad, A.; and Besold, T.
  2018.
\newblock Ultra-Strong Machine Learning: comprehensibility of programs learned
  with ILP.
\newblock \emph{Machine Learning}, 107: 1119–--1140.

\bibitem[{Qu et~al.(2021)Qu, Chen, Xhonneux, Bengio, and Tang}]{RNNLogic}
Qu, M.; Chen, J.; Xhonneux, L.-P.; Bengio, Y.; and Tang, J. 2021.
\newblock RNNLogic: Learning Logic Rules for Reasoning on Knowledge Graphs.
\newblock In \emph{ICLR 2021}.

\bibitem[{Richardson and Domingos(2006)}]{rich-dom}
Richardson, M.; and Domingos, P. 2006.
\newblock Markov logic networks.
\newblock \emph{Machine Learning}, 62: 107--136.

\bibitem[{Rochst\"atel and Riedel(2017)}]{roch}
Rochst\"atel, T.; and Riedel, S. 2017.
\newblock End-to-end differential proving.
\newblock In \emph{Neurips 2017}.

\bibitem[{Rudin et~al.(2022)Rudin, Chen, Chen, Huang, Semenova, and
  Zhong}]{interp-cynthia}
Rudin, C.; Chen, C.; Chen, Z.; Huang, H.; Semenova, L.; and Zhong, C. 2022.
\newblock Interpretable machine learning: Fundamental principles and 10 grand
  challenges.
\newblock \emph{Statistics Surveys}, 16: 1 -- 85.

\bibitem[{Sadeghian et~al.(2019)Sadeghian, Armandpour, Ding, and
  Wang}]{sadeghian19}
Sadeghian, A.; Armandpour, M.; Ding, P.; and Wang, D.~Z. 2019.
\newblock {DRUM}: End-To-End Differentiable Rule Mining On Knowledge Graphs.
\newblock In \emph{NeurIPS}.

\bibitem[{Schmid et~al.(2017)Schmid, Zeller, Besold, Tamaddoni-Nezhad, and
  Muggleton}]{mug1}
Schmid, U.; Zeller, C.; Besold, T.; Tamaddoni-Nezhad, A.; and Muggleton, S.~H.
  2017.
\newblock How does predicate invention affect human comprehensibility?
\newblock \emph{ILP 2016}, 52--67.

\bibitem[{Sen et~al.(2022{\natexlab{a}})Sen, Carvalho, Riegel, and
  Gray}]{Sen_Carvalho_Riegel_Gray_2022}
Sen, P.; Carvalho, B. W. S. R.~d.; Riegel, R.; and Gray, A. 2022{\natexlab{a}}.
\newblock Neuro-Symbolic Inductive Logic Programming with Logical Neural
  Networks.
\newblock \emph{Proceedings of the AAAI Conference on Artificial Intelligence},
  36(8): 8212--8219.

\bibitem[{Sen et~al.(2022{\natexlab{b}})Sen, de~Carvalho, Abdelaziz,
  Kapanipathi, Roukos, and Gray}]{sen_et_al}
Sen, P.; de~Carvalho, B. W. S.~R.; Abdelaziz, I.; Kapanipathi, P.; Roukos, S.;
  and Gray, A. 2022{\natexlab{b}}.
\newblock Logical Neural Networks for Knowledge Base Completion with Embeddings
  \& Rules.
\newblock \emph{The 2022 Conference on Empirical Methods in Natural Language
  Processing}.

\bibitem[{Shen et~al.(2018)Shen, Chen, Huang, Guo, and
  Gao}]{NEURIPS2018_c6f798b8}
Shen, Y.; Chen, J.; Huang, P.-S.; Guo, Y.; and Gao, J. 2018.
\newblock M-Walk: Learning to Walk over Graphs using Monte Carlo Tree Search.
\newblock In \emph{Neurips 2018}.

\bibitem[{Sun et~al.(2020)Sun, Vashishth, Sanyal, Talukdar, and Yang}]{sun}
Sun, Z.; Vashishth, S.; Sanyal, S.; Talukdar, P.; and Yang, Y. 2020.
\newblock A Re-evaluation of Knowledge Graph Completion Methods.
\newblock In \emph{Proceedings of the 58th Annual Meeting of the Association
  for Computational Linguistics}, 5516--5522.

\bibitem[{Teru, Denis, and Hamilton(2020{\natexlab{a}})}]{teru}
Teru, K.; Denis, E.; and Hamilton, W. 2020{\natexlab{a}}.
\newblock Inductive Relation Prediction by Subgraph Reasoning.
\newblock \emph{ICML 2020}.

\bibitem[{Teru, Denis, and Hamilton(2020{\natexlab{b}})}]{grail}
Teru, K.~K.; Denis, E.~G.; and Hamilton, W.~L. 2020{\natexlab{b}}.
\newblock Inductive relation prediction by subgraph reasoning.
\newblock \emph{ICML 2020}.

\bibitem[{Toutanova and Chen(2015)}]{toutanova}
Toutanova, K.; and Chen, D. 2015.
\newblock Observed Versus Latent Features for Knowledge Base and Text
  Inference.
\newblock In \emph{Proceedings of the 3rd Workshop on Continuous Vector Space
  Models and their Compositionality (CVSC)}, 57--66.

\bibitem[{Wang et~al.(2017{\natexlab{a}})Wang, Mao, Wang, and Guo}]{emb-survey}
Wang, Q.; Mao, Z.; Wang, B.; and Guo, L. 2017{\natexlab{a}}.
\newblock Knowledge graph embedding: A survey of approaches and applications.
\newblock \emph{IEEE TKDE}, 29: 2724--2743.

\bibitem[{Wang et~al.(2017{\natexlab{b}})Wang, Rudin, Doshi-Velez, Liu,
  Klampfl, and MacNeille}]{rudin17}
Wang, T.; Rudin, C.; Doshi-Velez, F.; Liu, Y.; Klampfl, E.; and MacNeille, P.
  2017{\natexlab{b}}.
\newblock A Bayesian framework for learning rule sets for interpretable
  classification.
\newblock \emph{Journal of Machine Learning Research}, 18: 1--–37.

\bibitem[{Xiong, Hoang, and Wang(2017)}]{xiong}
Xiong, W.; Hoang, T.; and Wang, W.~Y. 2017.
\newblock Deeppath: a reinforcement learning method for knowledge graph
  reasoning.
\newblock In \emph{EMNLP 2017}.

\bibitem[{Yang, Yang, and Cohen(2017)}]{yang17}
Yang, F.; Yang, Z.; and Cohen, W.~W. 2017.
\newblock Differentiable Learning of Logical Rules for Knowledge Base
  Reasoning.
\newblock In \emph{Advances in Neural Information Processing Systems 30}.

\bibitem[{Yang and Song(2020)}]{yang-song}
Yang, Y.; and Song, L. 2020.
\newblock Learn to explain efficiently via neural logic inductive learning.
\newblock \emph{ICLR 2020}.

\end{thebibliography}

\appendix
\section{Appendix}

\section{Proof of Theorem 1}

\begin{proof}
Let $(\bar \eta, \bar w)$ be an optimal solution of IPR. By definition, $\bar w$ has $|\mathcal{K}|$ components, and
$\bar \eta$ has $m = |E_r|$ components, all of which are binary, because of the form of the objective function.
Let $C_k$ be the clause associated with rule $k$. By definition, we have $a_{ik} = C_k(X_i, Y_i)$. 
Consider the function 
\begin{equation*} 
f(X,Y) = \vee_{k : \bar w_k = 1} C_k(X,Y) = \vee_{k \in \mathcal{K}} \bar w_k C_k(X,Y).
\end{equation*}
Therefore, $f : V \times V \rightarrow \{0,1\}$.
As $(\bar\eta, \bar w)$ satisfies equation (4), we have
\[ \sum_{k \in \mathcal{K}} a_{ik}\bar w_k + \bar\eta_i \geq 1 \mbox{ for all } i \in E_r.\]
Therefore, either $\eta_i = 0$ and $\sum_{k \in \K} a_{ik}\bar w_k \geq 1$, or $\eta_i = 1$ and $\sum_{k \in \K} a_{ik}\bar w_k  = 0$.
We can see that $f(X_i, Y_i) = 1$ if and only if $\sum_{k \in \K} a_{ik}\bar w_k \geq 1$, and therefore
\[ f(X_i, Y_i) + \bar\eta_i \geq 1 \mbox{ for all } i\in E_r.\]
Therefore, either $f(X_i, Y_i) = 1$ or $\bar \eta_i = 1$. Furthermore, $f(X,Y)$ is a function for which fewest number of values $\bar\eta_i$ are 1 or the highest number of values $\bar\eta_i$ are 0, as $(\bar \eta, \bar w)$ form an optimal solution of IPR. In other words, $f(X,Y)$ "covers" the largest number of edges of $E_r$ (covering means $f(X_i, Y_i) = 1$) among all possible functions that can be formed as a disjunction of rule clauses with complexity at most $\kappa$.
For each $i$ such that $\bar \eta_i = 0$, we have
\[ f(X_i, Y_i) = 1  \mbox{ and } f(X_i, Z) \leq 1 \mbox{ and } f(Z,Y_i) \leq 1\] for all $Z \in V$ as $f(X,Y) = 0$ or 1 for any entities $X,Y \in V$.
If we take the facts in the training set as a test set, and use $f(X,Y)$ as a scoring function and use {\em optimistic} ranking of scores, then for each $i$ such that $f(X_i, Y_i) = 1$, we have  $f(X_i, Y_i) \geq f(X_i, Z)$ and $f(X_i, Y_i) \geq f(Z,Y_i)$ for all entities $Z$. Therefore the rank of $Y_i$ among all entities $Z$ while scoring $(X_i, r, Z)$ (denoted by $rr_i$)  is 1, and the rank of $X_i$ among all the entities $Z$ while scoring $(Z, r, Y_i)$ (denoted by $lr_i$) is 1. On the other hand $rr_i \geq 1$ and $lr_i \geq 1$ if $f(X_i,Y_i) = 0$. Therefore $1/rr_i \geq 1-\bar \eta_i$ and $1/lr_i \geq 1-\bar\eta_i$. The MRR of the prediction function is
\begin{eqnarray*} & (\sum_{i=1}^m \frac{1}{rr_i} + \sum_{i=1}^m \frac{1}{lr_i})/2m & \geq 2(m - \sum_{i=1}^m \bar \eta_i) / 2m \nonumber\\
&& = 1 - \frac{1}{m}\sum_{i=1}^m \bar\eta_i.
\end{eqnarray*}
But $\sum_{i=1}^m \bar\eta_i$ is the optimal objective function value.
Thus a lower value of $\sum_{i=1}^m\eta_i$ yields a better lower bound on the MRR computed via optimistic ranking.
\end{proof}


\section{Additional experimental details}
\label{additional_exp_details}

Table \ref{table_tau} contains the list of values of the parameter $\tau$ given as input for each dataset. 
For larger datasets, this hyperparameter search is time-consuming, which is why we use fewer candidate $\tau$ values.

\begin{table}[h]
\centering
\resizebox{\linewidth}{!}{
\begin{tabular}{ll}  
\toprule
Datasets & Values of $\tau$ \\
\midrule
Kinship	   & 0.02, 0.025, 0.03, 0.035, 0.04, 0.045, 0.05, 0.055, 0.06  \\[5pt]
UMLS       & 0.02, 0.03, 0.04, 0.05, 0.0055, 0.06, 0.07, 0.08, 0.09, 0.1   \\[5pt]
WN18RR	   & 0.0025, 0.003, 0.0035, 0.004, 0.0045 \\[5pt]
FB15k-237  & 0.005, 0.01, 0.025, 0.05, 0.1, 0.25 \\[5pt]
YAGO3-10   & 0.005, 0.01, 0.03, 0.05, 0.07 \\[5pt]
\bottomrule 
\end{tabular}
}
\caption{Values of the parameter $\tau$ for each dataset.}
\label{table_tau}
\end{table}

Given the lists of values in Table \ref{table_tau}, removing values from this list reduces the MRR, but not too much. But if we use entirely different values, the MRR can drop significantly.

We use coarse-grained parallelism in our code.
For example, FB15K-237 has 237 relations, and we run the rule learning problem for each relation on a different thread. As we only have 60 cores, multiple threads are assigned to the same core by the operating system.

The run times increase significantly with increasing number of facts, and with increasing edge density in the knowledge graph. Recall that the rule learning linear program  (LPR) that we solve for each relation $r$ has a number of constraints equal to $|E_r|+1$, the number of edges labeled by relation $r$ in the knowledge graph. Given candidate rule $k$ in LPR, we need to compute $a_{ik} = C_k(X_i,Y_i)$ for each edge $i$ in $E_r$, where the $i$th edge in $E_r$ is $X_i \overset{r}{\rightarrow} Y_i$. Computing $a_{ik}$ increases superlinearly with increasing average node degree and increasing path length, and so does $\textup{neg}_k$ (both calculations involve an operation similar to a BFS or DFS).
The cost of solving a linear program (LP) grows superlinearly with the number of constraints. The larger datasets (WN18RR, FB15K-237, and YAGO3-10) all have some relation which has many more associated facts/edges than the average number per relation. This leads to significant run times, both in setting up LPR for the relation and in solving LPR, in the case of FB15K-237 and YAGO3-10, and also to reduced benefits of parallelism. For example, for YAGO3-10, the relations {\it isAffiliatedTo} and {\it playsFor} have 373,783 and 321,024 associated facts, respectively, out of a total of roughly a million facts. Most relations are completed in a short amount of time, while these two relations run for a long time on a single thread each. A natural approach to reducing the run time would be to sample some of these facts while setting up LPR, but we do not do this in this work. For YAGO3-10, we do not compute $\textup{neg}_k$ exactly and use sampling to obtain an approximation, as described in the main document.

One can easily parallelize the hyperparameter search process in our code.
Other operations which can be parallelized are the computation of the coefficients $a_{ik}$ in LPR, but we have not done so. We thus believe there is scope for reducing our run times even further. We note that the operations we perform are not well-suited to run on GPUs.

\section{Additional code details}

For the experiments in this paper, in Rule Heuristic 1, we simply generate all length one and length two rules that create a relational path from the tail to head node for at least one edge $(X_i, Y_i)$ associated with relation $r$, while creating LPR for relation $r$.

In Rule Heuristic 2, for every edge $X \overset{r}{\rightarrow} Y$ in $E_r$ (associated with relation $r$), we find a shortest path, using breadth-first search, from $X_i$ to $Y_i$ in the knowledge graph $\mathcal{G}$ that does not use the edge $X \overset{r}{\rightarrow} Y$. However, when we perform column generation, we do not find a shortest path between $X_i$ and $Y_i$ for every directed edge in $E_r$. Instead, we only consider edges $i \in E_r$ that are not "implied" by the currently chosen weighted combination of rules. Such edges are indicated by large dual values, as discussed earlier. A natural improvement to this algorithm would be to find rules which create relational paths between multiple pairs of tail and head nodes $X_i$ and $Y_i$ which have large dual values.

During the search for the best $\tau$ abd $\kappa$ values, we first set up LPR or LPR$_i$ (for some $i$, when we do column generation) for a fixed value of $\tau$ and $\kappa$. Subsequently, we do not add any more rules/columns, and instead change the values of $\tau$ and $\kappa$, and evaluate the resulting LP solution on a validation set by computing the MRR, and then choosing the $\tau, \kappa$ combination which gives the best value of MRR on the validation set.
During training, we obtain a weighted combination of rules for each relation separately, and then evaluate all these relations on the test set after training is complete. 

All relational paths that we create (either in shortest-path calculations or during evaluation on the test set) are {\em simple}, i.e., they do not repeat nodes.

All our code is written in C++. The LP Solver we use is IBM CPLEX \citep{cplex}, which is a commercial MIP solver (though available for free for academic use). Any high-quality LP solver can be used instead of CPLEX, though the interface functions in our code which have CPLEX-specific functions calls would need to be changed.

\section{Using rules from other methods}

In Tables~\ref{table:anyburl_plus_lprules} and~\ref{table:rnnlogic_plus_lprules} we show our results obtained using rules and rule weights generated by AnyBURL and RNNLogic, respectively, across the four scenarios discussed earlier.

\begin{table}[h]
\centering
\small
\resizebox{\linewidth}{!}{
\begin{tabular}{llcccc}  
\toprule
Problem   & Scenario & MRR & Hits@1 & Hits@10 & Avg \#rules \\
\midrule
Kinship   & A & 0.626 & 0.503 & 0.901 & 6653.1 \\
          & B & 0.525 & 0.385 & 0.856 & 6653.1 \\
          & C & 0.769 & 0.656 & 0.974 & 27.9   \\
          & D & 0.771 & 0.663 & 0.976 & 32.4   \\
UMLS      & A & 0.940 & 0.916 & 0.985 & 1837.6 \\
          & B & 0.797 & 0.740 & 0.943 & 1837.6 \\
          & C & 0.891 & 0.840 & 0.978 & 3.8    \\
          & D & 0.879 & 0.819 & 0.974 & 4.1    \\
FB15k-237 & A & 0.226 & 0.166 & 0.387 & 79.9   \\
          & B & 0.267 & 0.190 & 0.424 & 79.9   \\
          & C & 0.247 & 0.167 & 0.411 & 9.7    \\
          & D & 0.260 & 0.179 & 0.427 & 16.8   \\
WN18RR    & A & 0.454 & 0.423 & 0.527 & 47.3   \\
          & B & 0.392 & 0.332 & 0.515 & 47.3   \\
          & C & 0.360 & 0.295 & 0.497 & 9.8    \\
          & D & 0.461 & 0.425 & 0.532 & 15.8   \\
YAGO3-10  & A & 0.449 & 0.381 & 0.598 & 63.0   \\
          & B & 0.480 & 0.401 & 0.627 & 63.0   \\
          & C & 0.433 & 0.347 & 0.592 & 9.0    \\
          & D & 0.433 & 0.347 & 0.594 & 9.9    \\
\bottomrule
\end{tabular}
}
\caption{Combining rules from AnyBURL with our LP formulation and rules.}
\label{table:anyburl_plus_lprules}
\end{table}

\begin{table}[h]
\centering
\small
\resizebox{\linewidth}{!}{
\begin{tabular}{llcccc}  
\toprule
Problem   & Scenario & MRR & Hits@1 & Hits@10 & Avg \#rules \\
\midrule
Kinship   & A & 0.621 & 0.557 & 0.922 & 400.0 \\
          & B & 0.535 & 0.385 & 0.878 & 400.0 \\
          & C & 0.557 & 0.384 & 0.935 & 12.0   \\
          & D & 0.754 & 0.641 & 0.969 & 21.1   \\
UMLS      & A & 0.703 & 0.626 & 0.930 & 200.0 \\
          & B & 0.787 & 0.725 & 0.933 & 200.0 \\
          & C & 0.753 & 0.663 & 0.958 & 2.8    \\
          & D & 0.862 & 0.805 & 0.974 & 4.1    \\
WN18RR    & A & 0.423 & 0.407 & 0.514 & 399.3   \\
          & B & 0.293 & 0.215 & 0.471 & 399.3   \\
          & C & 0.362 & 0.295 & 0.493 & 13.4    \\
          & D & 0.461 & 0.424 & 0.533 & 14.5   \\
\bottomrule
\end{tabular}
}
\caption{Combining rules from RNNLogic with our LP formulation and rules.}
\label{table:rnnlogic_plus_lprules}
\end{table}

\section{Analysis of YAGO3-10}

\begin{table*}[htbp]
\centering
\small
\begin{tabular}{lllc}  
\toprule
Relation & Weight & Rule & MRR \\
\midrule
{\it isAffiliatedTo(x,y)} & 1 & {\it playsFor(x,y)} & 0.463 \\[5pt]
                    & 1 & {\it isAffiliatedTo(x,a) $ \wedge $ R\_isAffiliatedTo(a,b) $\wedge$ isAffiliatedTo(b,y)} & 0.582 \\[5pt]
                    & 1 & {\it graduatedFrom(x,a) $ \wedge $ R\_graduatedFrom(a,b) $\wedge$ isAffiliatedTo(b,y)} &  \\[5pt]
                    & 1 & {\it isPoliticianOf(x,a) $ \wedge $ R\_isPoliticianOf(a,b) $\wedge$ isAffiliatedTo(b,y)} &  \\[5pt]
                    & 0.5 & {\it livesIn(x,a) $ \wedge $ R\_livesIn(a,b) $\wedge$ isAffiliatedTo(b,y)} &  0.585 \\[5pt]
\midrule
{\it playsFor(x,y)}       & 1 & {\it isAffiliatedTo(x,y)} & 0.504 \\[5pt]
                    & 1 & {\it playsFor(x,a) $ \wedge $ R\_isAffiliatedTo(a,b) $\wedge$ playsFor(b,y)} & 0.561 \\[5pt]
\bottomrule 
\end{tabular}
\caption{Rules generated by LPRules for two relations in YAGO3-10. The MRR values for a particular rule were calculated using only the rules in the same line or above.}
\label{table_s20}
\end{table*}

We obtained an MRR of 0.449 with LPRules. We next analyze our performance and observe that it is primarily due to the perfomance on two relations, namely {\it IsAffiliatedTo} and {\it playsFor}, which together account for 64.4\% of the facts in the training set, and 63.3\% of the facts in the test set.
In Table~\ref{table_s20}, we provide the rules generated by LPRules for these two relations.
Here {\it R\_isAffiliatedTo} is the reverse of the relation {\it isAffiliatedTo} (and denoted by {\it isAffiliatedTo$^{-1}$} in the main document).
The rule and weight columns together give the weighted combination of rules generated for a relation. The MRR column gives the MRR value that would be generated if the test set consisted only of facts associated with the relation in the "Relation" column, and the weighted combination of rules consisted only of the rules in the same line or above.
Thus we can see that if we took only the first two rules for {\it isAffiliatedTo}, and the rules for {\it playsFor}, then the MRR would be at least 0.56 on the test set facts associated with these two relations.
As these two relations account for 63.3\% of the test set, just the four rules mentioned above would yield an MRR $ \geq 0.56 \times 0.633 \approx 0.354$, as opposed to the MRR of 0.449 that we obtained. 

The information in the table suggests a direct correlation between the relations {\it isAffiliatedTo} and {\it playsFor}. Indeed, we verify that for about 75\% of the facts {\it (x, isAffiliatedTo, y)}, we also have {\it (x, playsFor, y)} as a fact in the training set. Similarly, 87\% of the {\it playsFor} facts are explained by the isAffiliatedTo relation in the training set.

A natural question is the following: Is the second rule for {\it isAffiliatedTo} a ``degenerate" rule and does it simply reduce to {\it isAffiliatedTo} because the entity {\it b} is the same as entity {\it a} when we traverse a relational path from {\it x} to {\it y} in the training KG.
To give an example that this is not the case, consider the following fact in the test set: {\it (Pablo\_Bonells, isAffiliatedTo, Club\_Celaya)}. In the training data, the following three facts imply the previous fact by application of the second rule:
{\it (Pablo\_Bonells, isAffiliatedTo, Club\_León)}, {\it (Salvador\_Luis\_Reyes, isAffiliatedTo, Club\_León)}, and {\it (Salvador\_Luis\_Reyes, isAffiliatedTo, Club\_Celaya)}. The first rule is also applicable as the training data contains the fact {\it (Pablo\_Bonells, playsFor, Club\_Celaya)}.
We have similarly verified that the second rule for {\it playsFor} creates nontrivial relational paths, where the nodes are not repeated.

\end{document}